\documentclass{article}

\usepackage{microtype}
\usepackage{graphicx}
\usepackage{subfigure}
\usepackage{booktabs} 

\usepackage{hyperref}

\usepackage{graphicx} 
\usepackage{subfigure}
\usepackage{amsmath}
\usepackage{bm}
\usepackage{multirow}
\usepackage{amssymb}
\usepackage{subfigure}
\usepackage{wrapfig}

\usepackage[accepted]{icml2020}

\icmltitlerunning{Deep Multi-Task Augmented Feature Learning via Hierarchical Graph Neural Network}

\begin{document}

\twocolumn[
\icmltitle{Deep Multi-Task Augmented Feature Learning via Hierarchical Graph Neural Network}



\icmlsetsymbol{equal}{*}

\begin{icmlauthorlist}
\icmlauthor{Pengxin Guo}{sustech}
\icmlauthor{Chang Deng}{sustech}
\icmlauthor{Linjie Xu}{sustech}
\icmlauthor{Xiaonan Huang}{sustech}
\icmlauthor{Yu Zhang}{sustech}
\end{icmlauthorlist}

\icmlaffiliation{sustech}{Department of Computer Science and Engineering, Southern University of Science and Technology, China}

\icmlcorrespondingauthor{Yu Zhang}{yu.zhang.ust@gmail.com}

\icmlkeywords{Machine Learning, ICML}

\vskip 0.3in
]



\printAffiliationsAndNotice

\begin{abstract}

Deep multi-task learning attracts much attention in recent years as it achieves good performance in many applications. Feature learning is important to deep multi-task learning for sharing common information among tasks. In this paper, we propose a Hierarchical Graph Neural Network (HGNN) to learn augmented features for deep multi-task learning. The HGNN consists of two-level graph neural networks. In the low level, an intra-task graph neural network is responsible of learning a powerful representation for each data point in a task by aggregating its neighbors. Based on the learned representation, a task embedding can be generated for each task in a similar way to max pooling. In the second level, an inter-task graph neural network updates task embeddings of all the tasks based on the attention mechanism to model task relations. Then the task embedding of one task is used to augment the feature representation of data points in this task. Moreover, for classification tasks, an inter-class graph neural network is introduced to conduct similar operations on a finer granularity, i.e., the class level, to generate class embeddings for each class in all the tasks use class embeddings to augment the feature representation. The proposed feature augmentation strategy can be used in many deep multi-task learning models. we analyze the HGNN in terms of training and generalization losses. Experiments on real-world datastes show the significant performance improvement when using this strategy.

\end{abstract}

\newtheorem{theorem}{Theorem}
\newtheorem{lemma}{Lemma}
\newtheorem{definition}{Definition}
\newtheorem{remark}{Remark}
\newtheorem{corollary}{Corollary}

\section{Introduction}

Multi-task learning \cite{caruana97,zy17b} aims to leverage useful information contained in multiple learning tasks to improve their performance simultaneously. During past decades, many multi-task learning models have been proposed to identify the shared information which can take a form of the instance, feature, and model, leading to three categories including instance-based multi-task learning \cite{bbls08}, feature-based multi-task learning \cite{aep06,otj06,lpz09,zyx10,ls12,shinohara16,msgh16,ljd19}, and model-based multi-task learning \cite{az05,jbv08,bcw07,zy10a,jrsr10,kd12,hz15a,hz16,zwy18}.


For multi-task learning, it is important to decide how to represent a task and how to leverage the relationship between tasks to improve the performance of all the tasks based on the task representation. For the first issue, multi-task learning can use the instance, feature or parameter as a media to construct the task representation. Built on the solution to the first issue, the task relationship can be learned via the chosen media, leading to a classification of multi-task learning including instance-based multi-task learning, feature-based multi-task learning, and parameter-based multi-task learning.

Different from existing studies, we think that the training dataset of a task contains important information to determine the task representation in the first issue and we can extract information from the training dataset as the task representation. However, it is not easy to derive a representation from a dataset based on convolutional neural networks or recurrent neural networks. Our solution is to represent a dataset as a graph where nodes represent data points and edges denote the similarities between data points. For the second issue, the task relationship can also be represented in a graph. So the two important issues can be related to graphs.


Inspired by this idea, in this paper, we propose a Hierarchical Graph Neural Network (HGNN) to further improve the performance of multi-task learning models by learning augmented features. The HGNN consists of two-level graph neural networks. In the first level, an intra-task graph neural network is to learn a powerful representation for each data point in a task by aggregating its neighbored data points in this task. Based on the representation learned in the first level, we can generate the task embedding, which is a representation for this task, in a way similar to max pooling. For classification tasks, we can generate the class embedding for each class in this task based on max pooling. Based on task embeddings of all the tasks generated in the first level, an inter-task graph neural network in the second level updates all the task embeddings based on the attention mechanism. For classification tasks, an inter-class graph neural network is introduced in the second level to update all the class embeddings based on neighbored class embeddings. Finally, each of the learned task embeddings as well as the class embeddings for classification tasks is used to augment the feature representation of all the data points in the corresponding task. The proposed HGNN can be used in many multi-task learning models. We analyze the use of HGNN in terms of both the training loss and generalization loss. Extensive experiments show the effectiveness of the proposed HGNN.

\section{Related Works}

\citet{lfdqc18} explore the problem of learning the relationship between multiple tasks dynamically and formulate this problem as a message passing process over a graph neural network. 
\citet{makfs18} solve relative attribute learning via a message passing scheme on a graph and the main idea is that relative attribute learning naturally benefits from exploiting the dependency graph among different relative attributes of images. The multi-task attention network proposed in \cite{ljd19} consists of a single shared network containing a global feature pool and a soft-attention module for each task that allows to learn task-specific feature-level attentions. The soft-attention module can learn both task-specific features from global features and shared features across different tasks. \citet{lhyg19} present a graph star net which utilizes the message-passing and attention mechanisms for multiple prediction tasks, including node classification, graph classification, and link prediction. Even though the aforementioned works propose GNN or the attention mechanism for multi-task learning, none of them use a hierarchical version of GNN as well as the attention mechanism to learn augmented features for multi-task learning, which is the focus of this paper.

\begin{figure*}[!htbp]
\vskip -0.1in
\centering
\includegraphics[scale=0.75]{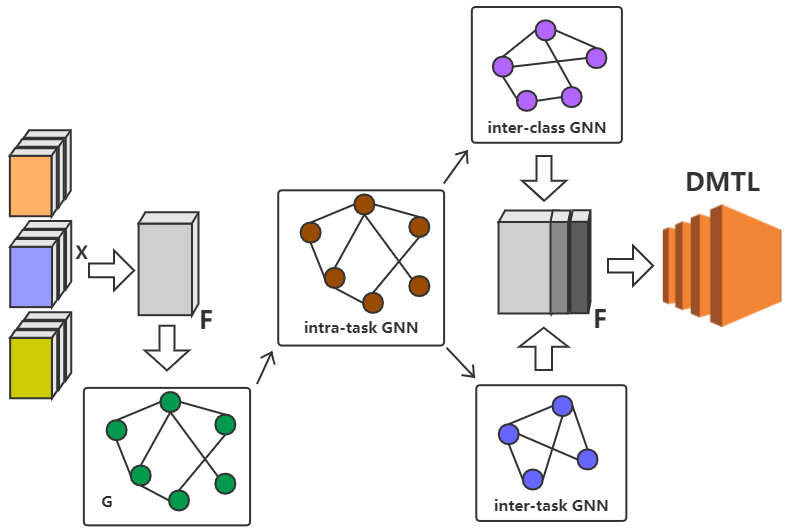}
\vskip -0.1in
\caption{An illustration of the hierarchical graph neural network for multi-task learning.}
\label{HGNN}
\vskip -0.1in
\end{figure*}

\citet{ksjlkk19} propose a hierarchical attention network for stock prediction which can selectively aggregate information on different relation types and add the information to each representation of the company for the stock market prediction. By considering the market index as an entire graph and constituent companies as individual nodes, this method is used for predicting not only individual stock prices but also market index movements, which is similar to the graph classification task. However, after obtaining the additional information to each representation, it only uses the neighbored nodes but not all the nodes in the graph to aggregate the information, hence it cannot get whole information of the graph. Moreover, this method adds the additional information to the original feature representation, which is different from the concatenation method used in this paper.  \citet{rsp19} propose a Hierarchical graph Attention-based Multi-Agent actor-critic (HAMA) method, which employs a hierarchical graph neural network to effectively model the inter-agent relationships in each group of agents and inter-group relationships among groups, and additionally employ inter-agent and inter-group attentions to adaptively extract state-dependent relationships among agents. However, similar to \cite{ksjlkk19}, the HAMA method, a network stacking multiple Graph Attention Networks (GAT) hierarchically, only processes local observations of each agent but not all the information in the graph, which cannot capture enough information. Different from these two methods, we first use an intra-task graph neural network to generate a task embedding for each task by using all the data in this task. By using all the data in a task, we can get a more accurate and robust task embedding. Then, an inter-task graph neural network is used to update task embeddings of all the tasks based on the inter-task structure, which makes each task embedding contains useful information from embeddings of other tasks. The class embeddings can be obtained in a similar way. So the task embeddings can represent the relationship of all the tasks and the class embeddings can represent the relationship of all the classes in all the tasks. To the best of our knowledge, we are the first to use the feature augmentation strategy in multi-task learning.

\section{Hierarchical Graph Neural Network}

In this section, we introduce the proposed architecture, the Hierarchical Graph Neural Network (HGNN), for deep multi-task learning. Whilst the architecture can be incorporated into any multi-task learning network, in the following sections we show how to build the HGNN upon a multi-task network.

\subsection{Overview of The Architecture}

The HGNN consists of two-level GNNs. The first-level GNN is an intra-task GNN to aggregate all the information contained in the data of a task to generate a task representation, which is called the task embedding. In the second level, based on the generated task embeddings in the first level, an inter-task GNN is used to update all the task embeddings by sharing information among all the tasks. Finally the task embeddings are used to augment the feature representation of the data to improve the learning performance. For classification tasks, we can learn augmented features in a fine granularity - the class level. That is, another intra-task GNN is used to aggregate all the information in a class of a task to generate a class embedding. Then based on class embeddings in all the tasks, another inter-class GNN is used to update them. Finally, both task embeddings and class embeddings are used to augment the feature representation. The whole architecture of HGNN is shown in Figure \ref{HGNN}.

\subsection{The Model}

Suppose that there are $m$ multi-class classification tasks where each task has $k$ classes. The training dataset of the $i$th task consists of $n_i$ pairs of data samples and corresponding labels, i.e., $\mathcal{D}_i=\{(\mathbf{x}^i_j,y^i_j)\}_{j=1}^{n_i}$ where $y^i_j\in\{1,\ldots,k\}$.

For $\mathbf{x}^i_j$, we first define its hidden representation as
\begin{equation}
\hat{\mathbf{h}}^i_j=\sigma_s(\hat{\mathbf{W}}_s\mathbf{x}^i_j+\hat{\mathbf{b}}_s),\label{equ_transform}
\end{equation}
where $\sigma_s(\cdot)$ can be any activation function such as the ReLU function, and $\hat{\mathbf{W}}_s,\hat{\mathbf{b}}_s$ are shared parameter among all the tasks.

For the intra-task GNN, we first construct an adjacency matrix $\mathbf{G}_i$ for the $i$th task based on the hidden representation and label information. Specifically, the $(j,l)$th entry in $\mathbf{G}_i$, $g^i_{jl}$, can be defined as
\begin{equation*}
g^i_{jl}=\left\{ \begin{array}{ll}
\exp\{-\|\hat{\mathbf{h}}^i_j-\hat{\mathbf{h}}^i_l\|_2^2\} & \textrm{if $y^i_j=y^i_l$}\\
-\exp\{-\|\hat{\mathbf{h}}^i_j-\hat{\mathbf{h}}^i_l\|_2^2\} & \textrm{otherwise}
\end{array} \right.,
\end{equation*}
where $\|\cdot\|_2$ denotes the $\ell_2$ norm of a vector. Then the intra-task GNN can be defined as
\begin{equation}
\mathbf{H}_i=\sigma_h(\mathbf{W}_h^i\mathbf{X}_i+\hat{\mathbf{H}}_i\mathbf{G}_i+\mathbf{b}_h^i\mathbf{1}),\label{equ_intra_GNN}
\end{equation}
where $\sigma_h(\cdot)$ can be any activation function such as the ReLU function, $\mathbf{X}_i=(\mathbf{x}^i_1, \ldots, \mathbf{x}^i_{n_i})$, $\hat{\mathbf{H}}_i=(\hat{\mathbf{h}}^i_1,\ldots,\hat{\mathbf{h}}^i_{n_i})$, $\mathbf{1}$ denotes a vector of all ones with an appropriate size, $\mathbf{W}_h$ and $\mathbf{b}_h$ are the parameters in the GNN. The $j$th column in $\mathbf{H}_i$ denoted by $\mathbf{h}^i_j$ is a new hidden representation for $\mathbf{x}^i_j$. 
$\mathbf{G}_i$ in Eq. (\ref{equ_intra_GNN}) can make similar data points in the same class have similar representations in $\mathbf{H}_i$ and dissimilar data points from different classes have dissimilar representations. The intra-task GNN can have two or more layers each of which is defined as in Eq. (\ref{equ_intra_GNN}).

Based on the intra-task GNN, the task embedding of the $i$th task is defined as
\begin{equation*}
\mathbf{e}_t^i=\max_l\{\mathbf{h}^i_l\},
\end{equation*}
where the max operation is conducted elementwisely. So the task embedding is obtained via the max pooling on all the data points in the $i$th task based on the hidden representation learned by the intra-task GNN. Similarly, the class embedding of the $j$th class in $i$th task is defined as
\begin{equation*}
\mathbf{e}_c^{i,j}=\max_{l:y^i_l=j}\{\mathbf{h}^{i}_l\},
\end{equation*}
which means that the class embedding of the $j$th class in the $i$th task is obtained via the max pooling on all the data points in the $j$th class of the $i$th task based on the intra-task GNN. We have tried other pooling methods such as the mean pooling but the performance is inferior to that of the max pooling. One reason is that the max pooling can bring some nonlinearity but the mean pooling is a linear operation.

The $m$ task embeddings $\{\mathbf{e}_t^i\}_{i=1}^m$ for the $m$ tasks can form a graph. The inter-task GNN is responsible of learning for the graph constructed by task embeddings $\{\mathbf{e}_t^i\}_{i=1}^m$ to generate new task embeddings $\{\hat{\mathbf{e}}_t^i\}_{i=1}^m$ by exchanging information among tasks. Here we use GAT as an implementation of the inter-task GNN.


In order to learn powerful task embeddings, each task embedding $\mathbf{e}_t^i$ is first transformed by a weight matrix $\mathbf{W}$. Then we perform \textit{self-attention} on the task embeddings. That is, an attentional mechanism computes \textit{attention coefficients} as
\begin{equation*}
d_{ij} = a(\mathbf{W}\mathbf{e}_t^i, \mathbf{We}_t^j)
\end{equation*}
where the attentional mechanism $a(\cdot, \cdot)$ we use is the cosine function, which is different from the original GAT. To make coefficients comparable across different task embeddings, we normalize them over $j$ by using the softmax function as
\begin{equation*}
\alpha_{ij} = \text{softmax}_j(d_{ij}) = \frac{\exp(d_{ij})}{\sum_{k} \exp(d_{ik})}.
\end{equation*}
Attention values can be viewed as a measure of task relations between each pair of tasks. Once obtained, the normalized attention coefficients are used as combination coefficients to compute a linear combination of the new task embeddings before potentially applying a nonlinear activation function $\sigma$ as
\begin{equation*}
\hat{\mathbf{e}}_t^i = \sigma\Big (  \sum_{j=1}^m \alpha_{ij} \mathbf{W} \mathbf{e}_t^j\Big ).
\end{equation*}
Based on this equation, we can see that $\hat{\mathbf{e}}_t^i$ contains useful information from embeddings of other tasks. In experiments, the inter-task GNN adopts two such layers to generate the new task embeddings.

Similarly, the $mk$ class embeddings $\{\mathbf{e}_c^{i,j}\}$ also can form a graph. We uses another inter-class GNN to generate new class embeddings $\{\hat{\mathbf{e}}_c^{i,j}\}$ in a way similar to task embeddings.

The learned task embeddings and class embeddings can be used to augment the data feature representation to form a more expressive one as $\tilde{\mathbf{h}}^i_l=(\mathbf{\hat{h}}^i_l,\hat{\mathbf{e}}_t^i, \hat{\mathbf{e}}_c^{i,j})$, where $(\cdot,\cdot,\cdot)$ denotes the concatenation operation. Then data in such augmented representation can be fed into a deep multi-task learning model to learn class labels.

\subsection{Testing Process}
At the testing process, we do not know the true label, hence we cannot directly concatenate the class embedding to the hidden representation. We use the following method to solve this problem. For each testing sample, we concatenate the class embedding of each class $c$ to the hidden representation $\hat{\mathbf{h}}^i_j$ as its new hidden representation and then compute the prediction probability that the testing sample belongs to class $c$ via the softmax function used in the multi-task neural network. Finally we choose class $c$ with the largest prediction probability as the predicted label. In mathematics, we predict the class label of a testing sample as
\begin{equation}
c^*=\arg\max_{l\in[k]}\mathbb{P}(y=l|f((\mathbf{\hat{h}}^i_*,\hat{\mathbf{e}}_t^i, \hat{\mathbf{e}}_c^{i,l}))),\label{equ_prediction_rule}
\end{equation}
where $[k]$ denotes a set of positive integers no larger than $k$, $\mathbf{\hat{h}}^i_*$ denotes the transformed testing sample before feeding into the HGNN as shown in Eq. (\ref{equ_transform}), and $f(\cdot)$ denotes the multi-task neural network used. Note that in the prediction rule (\ref{equ_prediction_rule}), the concatenated class embedding $\hat{\mathbf{e}}_c^{i,l}$ changes with $l$.

\subsection{Extension to Regression Tasks}
For the regression problems, there are only continuous labels and we cannot define class embeddings. So we only use task embeddings as the augmenter feature. Furthermore, the adjacency matrix $\mathbf{G}_i$ for the $i$th task is different from classification tasks. Specifically, the $(j,l)$th entry in $\mathbf{G}_i$, $g^i_{jl}$, for a regression task is defined as
\begin{equation*}
g^i_{jl}=\exp\{-\|\hat{\mathbf{h}}^i_j-\hat{\mathbf{h}}^i_l\|_2^2\}.
\end{equation*}
Since there is no class embedding, we do not need the prediction rule as in Eq. (\ref{equ_prediction_rule}).
The rest is identical to classification tasks.

\section{Analysis}

The proposed approach to augment the feature representation in HGNN is novel and here we provide some analyses to give insights into this model. For simplicity, we consider a linear single-task learning model by utilizing the task embedding only, which can provide insights for understanding deep multi-task learning models with task embeddings as well as class embeddings learned in HGNN.

The input space, which is a subset of a vector space, is denoted by $\mathcal{X}$ and the output space is denoted by $\mathcal{Y}$. Training samples $\{(\mathbf{x}_i,\mathbf{y}_i)_{i=1}^n\}\in \mathcal{X} \times \mathcal{Y}$ are distributed according to some unknown distribution $P$. Let $\ell:\mathbb{R}^{k}\times\mathcal{Y} \rightarrow \mathbb{R}^{+}$ be the loss function, where $k$ denotes the dimension of the label space. The learning function is defined as $f(\mathbf{x})=\mathbf{W}^\intercal\mathbf{x}$ where the superscript $^\intercal$ denotes the transpose and $\mathbf{W}$ is abused to denote the parameter in this linear learner. The expected loss is defined as $\mathcal{L}(\mathbf{W}) = \mathbb{E}[\ell(\mathbf{W}^\intercal\mathbf{x},\mathbf{y})]$. 
The empirical loss is defined as $\hat{\mathcal{L}}(\mathbf{W})=\frac{1}{n}\sum_{i=1}^n \ell (\mathbf{W}^\intercal \mathbf{x}_i,\mathbf{y}_i)$. 
The data matrix $\mathbf{X}$ is defined as $\mathbf{X}=(\mathbf{x}_1,\ldots,\mathbf{x}_n) \in \mathbb{R}^{p\times n}$ and the label matrix $\mathbf{Y}$ is defined as $\mathbf{Y}=(\mathbf{y}_1,\ldots,\mathbf{y}_n)\in\mathbb{R}^{k \times n}$. $\mathbf{e} \in \mathbb{R}^{q\times 1} $ denotes the task embedding and $\mathbf{E} =\mathbf{e}\mathbf{1}^\intercal \in \mathbb{R}^{q \times n}$ is the task embedding matrix for all the training data, where $\mathbf{1}$ denotes a column vector of all ones with an appropriate size.

Let us consider two models. The objective function of model 1 is formulated as
\begin{equation}
    \hat{\mathbf{W}}_1 = \underset{\mathbf{W_1}}{\operatorname{argmin}} \|\mathbf{Y}- \mathbf{W}_1^\intercal\mathbf{X} \|^2_2 +\lambda \|\mathbf{W_1} \|^2_2,\label{obj_analysis_model_1}
\end{equation}
and that of model 2 is
\begin{equation}
    \hat{\mathbf{W}}_2 = \underset{\mathbf{W_2}}{\operatorname{argmin}} \|\mathbf{Y}-\mathbf{W}_2^\intercal\hat{\mathbf{X}}  \|^2_2 +\lambda \|\mathbf{W_2} \|^2_2,\label{obj_analysis_model_2}
\end{equation}
where $\mathbf{W_1} \in \mathbb{R}^{p \times k}$, $\mathbf{W_2} \in \mathbb{R}^{(q+p) \times k}$, $\hat{\mathbf{x}}_i=(\mathbf{x}_i^\intercal,\mathbf{e}^\intercal)^\intercal$, $\hat{\mathbf{X}} = (\hat{\mathbf{x}}_1,\ldots,\hat{\mathbf{x}}_n) = (\mathbf{X}^\intercal,\mathbf{E}^\intercal)^\intercal \in \mathbb{R}^{(q+p)\times n}$, $\mathbf{I}$ is an identity matrix. So model 1 is a ridge regression model which can be applied to both classification and regression tasks and model 2 is a variant of model 1 with the task embedding incorporated. For training losses of those two models, we have the following result.\footnote{All the proofs can be found in the appendix.}

\begin{theorem}\label{theorem_training_loss_comparison}
If $\mathbf{X}$ and $\mathbf{E}$ satisfy $\mathbf{X}^\intercal\mathbf{X}\mathbf{E}^\intercal\mathbf{E}+\mathbf{E}^\intercal\mathbf{E}\mathbf{X}^\intercal\mathbf{X}+2\lambda \mathbf{E}^\intercal\mathbf{E} +\mathbf{E}^\intercal\mathbf{E}\mathbf{E}^\intercal\mathbf{E} \succeq 0$ where $\mathbf{M}_1\succeq\mathbf{M}_2$ means that $\mathbf{M}_1-\mathbf{M}_2$ is positive semidefinite, then the training loss of model 2 with the task embedding is always lower than that of model 1 without the task embedding. That is, we have
\begin{equation}
    \|\mathbf{Y}-\hat{\mathbf{W}}_1^\intercal\mathbf{X}\|_2^2\geq \|\mathbf{Y}-\hat{\mathbf{W}}_2^\intercal\hat{\mathbf{X}}\|_2^2
\end{equation}
\end{theorem}

\begin{remark}
Theorem \ref{theorem_training_loss_comparison} implies that for a model, incorporating the task embedding to augment the feature representation will incur a lower training loss than that without the task embedding. From the perspective of the model capacity, model 1 is a reduced version of model 2 by setting the task embedding to be zero and hence mode 2 has a larger capacity than model 1, making model 2 possess a large chance to have a lower training loss. The condition proposed in Theorem \ref{theorem_training_loss_comparison} is very easy to check and we can adjust $\lambda$ to ensure the positive semidefiniteness of the condition. 
\end{remark}

We also analyze the generalization bound of the two models. We first rewrite problems (\ref{obj_analysis_model_1}) and (\ref{obj_analysis_model_2}) into equivalent formulations as
\begin{align*}
\hat{\mathbf{W}}_1 = &\underset{ \|\mathbf{W_1} \|_2\leq W_* }{\operatorname{argmin}} \|\mathbf{Y}- \mathbf{W}_1^\intercal\mathbf{X} \|^2_2\\
\hat{\mathbf{W}}_2 = &\underset{ \|\mathbf{W_2} \|_2\leq W_* }{\operatorname{argmin}} \|\mathbf{Y}- \mathbf{W}_2^\intercal\hat{\mathbf{X}} \|^2_2.
\end{align*}
For the above two problems, we have the following result.

\begin{theorem}\label{theorem_generalization_bound}
Suppose $\left\|\mathbf{x}_i\right\|,\|\hat{\mathbf{x}}_i\| \leq X_*$, the task embedding satisfies the condition in Theorem \ref{theorem_training_loss_comparison}. Then for any $\delta>0$, with probability at least $1-\delta$, we have
{\small
\begin{align*}
&\mathbb{E}_{\mathbf{x},\mathbf{y}}(\|\mathbf{y}-\hat{\mathbf{W}}_1^\intercal\mathbf{x}\|_2^2) \\
&\leq \frac{1}{n}\sum_{i=1}^n\|\mathbf{y}_i-\hat{\mathbf{W}}_1^\intercal\mathbf{x}_i\|_2^2+ 4X_* \beta_*\sqrt{\frac{1}{n}}+2X_*\beta_*\sqrt{\frac{\log(1/\delta)}{2n}}\\
&\mathbb{E}_{\mathbf{x},\mathbf{y}}(\|\mathbf{y}-\hat{\mathbf{W}}_2^\intercal\hat{\mathbf{x}}\|_2^2) \\
&\leq \frac{1}{n}\sum_{i=1}^n\|\mathbf{y}_i-\hat{\mathbf{W}}_2^\intercal\hat{\mathbf{x}}_i\|_2^2+ 4X_* \beta_*\sqrt{\frac{1}{n}}+2X_*\beta_*\sqrt{\frac{\log(1/\delta)}{2n}}.
\end{align*}
}\noindent
\end{theorem}

\begin{remark}
According to Theorem \ref{theorem_generalization_bound}, the generalization upper-bound of model 2 with the use of the task embedding is lower than that without the task embedding because of the lower training loss of model 2 which has been proved in Theorem \ref{theorem_training_loss_comparison}. This may imply that there is a large chance that the expected loss of model 2 is lower than that of model 1, which can be verified in empirical studies in the next section.
\end{remark}

\section{Experiments}

In this section, we conduct empirical studies to test the performance of the proposed HGNN.



\subsection{Experimental Settings}

We conduct experiments on several benchmark datasets, including \textbf{ImageCLEF}, \textbf{Office-Caltech-10}, \textbf{Office-Home} and \textbf{SARCOS}.

The ImageCLEF dataset  is the benchmark for Image-
CLEF domain adaptation challenge which contains about
2,400 images from 12 common categories shared by four
tasks including \textit{Caltech-256} (\textbf{C}), \textit{ImageNet ILSVRC} (\textbf{I}), \textit{Pascal VOC 2012} (\textbf{P}), and \textit{Bing} (\textbf{B}). There are 50 images in each category and 600 images in each task.

The Office-Caltech-10 dataset includes 10 common categories shared by the Office-31 and Caltech-256 datasets. It contains four domains: \textit{Caltech} (\textbf{C}) that is sampled from Caltech-256 dataset, \textit{Amazon} (\textbf{A}) that contains images collected from the amazon website, \textit{Webcam} (\textbf{W}) and \textit{DSLR} (\textbf{D}) that are taken by the web camera and DSLR camera under the office environment. In our experiment, we regard each domain as a task.

The Office-Home dataset has 15,500 images across 65 classes in the office and home settings from four domains with a large domain discrepancy: \textit{Artistic images} (\textbf{Ar}), \textit{Clip art} (\textbf{Cl}), \textit{Product images} (\textbf{Pr}), and \textit{Real-world images} (\textbf{Rw}). In our experiment, we regard each domain as a task.

The SARCOS dataset studies a multi-output problem of learning the inverse dynamics of 7 SARCOS anthropomorphic robot arms, each of which corresponds to a task, based on 21 features, including seven joint positions, seven joint velocities, and seven joint accelerations. By following \cite{zy10a}, we randomly sample 2000 data points from each output to construct the dataset.

Since the proposed HGNN can be combined with many deep multi-task learning models as discussed before, we incorporate the HGNN into the Deep Multi-Task Learning (DMTL) which shares the first several layers as the common hidden feature representation for all the tasks as did in \cite{caruana97,zllt14,lmzcl15,zlzskyj15,mstgsvwy15,llc15}, Deep Multi-Task Representation Learning (DMTRL) \cite{yh17a}, and Trace Norm Regularised Deep Multi-Task Learning (TNRMTL) \cite{yh17b}, respectively, to show the benefit of the learned augmented features.

In experiments, we leverage the VGG-19 network pretrained on the ImageNet dataset as the backbone of the feature generator $G$. 
After that, all the multi-task learning models adopt a two-layer fully-connected architecture (\#data\_dim $\times$ 600 $\times$ \#classes) and the ReLU activation function is used. The first layer is shared by all tasks to learn a common representation and the second layer is for task-specific outputs. The HGNN learns 8-dimensional task embeddings and 8-dimensional class embeddings.

We use Adam with the learning rate varying as $\eta = \frac{0.02}{1 + p}$, where $p$ is the number of the iteration. By following GAT, We fix $F' = 8$ in experiments. We adopt mini-batch SGD with $\text{batch\_size} = 32$.  
Each experiment repeats for 5 times and we report the average performance as well as the standard deviation.

\subsection{Experimental Results}

\subsubsection{Results on Classification Tasks}

\begin{figure}[!htbp]
\centering
\includegraphics[width=8cm,height=5cm]{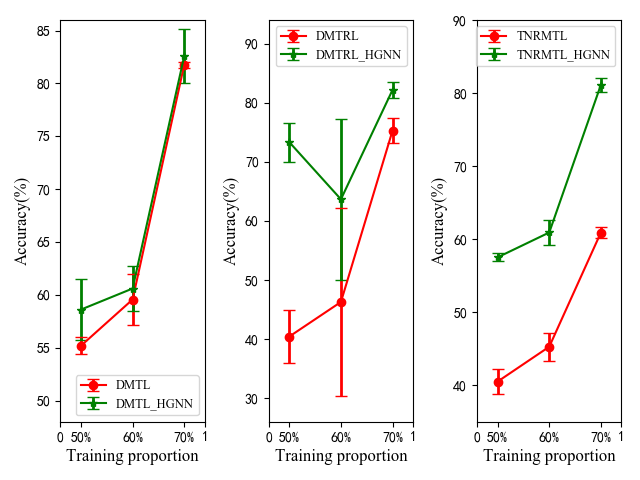}
\vskip -0.1in
\caption{Performance of different models on the ImageCLEF dataset when varying with the training proportion.}
\label{imageclef}
\vskip -0.1in
\end{figure}

\begin{figure}[!htbp]
\vskip -0.1in
\centering
\includegraphics[width=8cm,height=5cm]{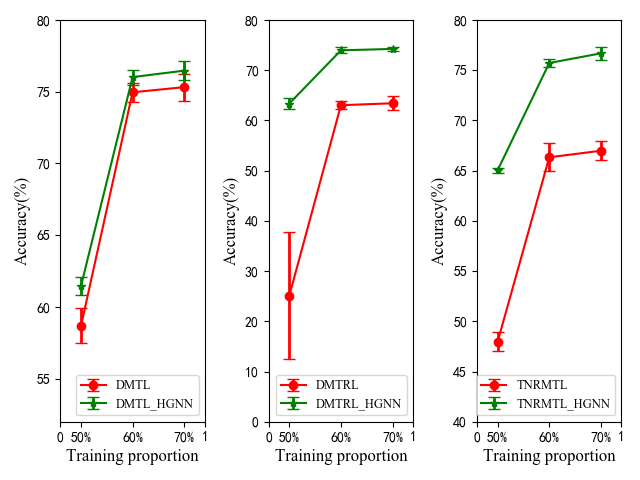}
\vskip -0.1in
\caption{Performance of different models on the Office-Home dataset when varying with the training proportion.}
\label{office_home}
\vskip -0.1in
\end{figure}

\begin{figure}[!ht]
\vskip -0.1in
\centering
\includegraphics[width=8cm,height=5cm]{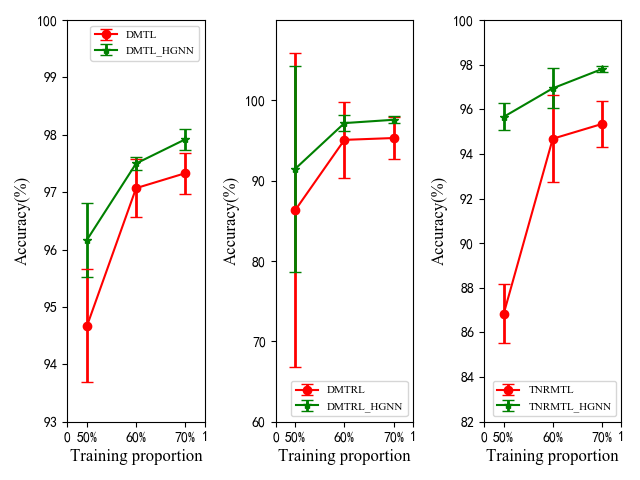}
\vskip -0.1in
\caption{Performance of different models on the Office-Caltech-10 dataset when varying with the training proportion.}
\label{office_caltech_10}
\vskip -0.1in
\end{figure}

For classification tasks, the performance measure is the classification accuracy. To investigate the effect of the size of the training dataset on the performance, we vary the proportion of training data from 50\% to 70\% at an interval of 10\% and plot the average test accuracy of different methods in Figures \ref{imageclef}-\ref{office_caltech_10}. According to results reported in these figures, we can see that the incorporation of the HGNN into baseline models improves the classification accuracy of all baseline models especially when the training proportion is small. 
As reported in Figures \ref{office_home} and \ref{office_caltech_10}, the incorporation of the HGNN boosts the performance of all the baseline on the Office-Caltech-10 and Office-Home datasets. For the DMTRL and TNRMTL models, the improvement is significant with the use of the HGNN. Moreover, when using augmented features learned by the HGNN, the standard deviation becomes smaller than the corresponding baseline model without using the HGNN under every experimental setting, which implies that the HGNN can improve the stability of baseline models to some extent. 

\subsubsection{Results on Regression Tasks}

\begin{figure}[!htbp]
\vskip -0.1in
\centering
\includegraphics[scale=0.33]{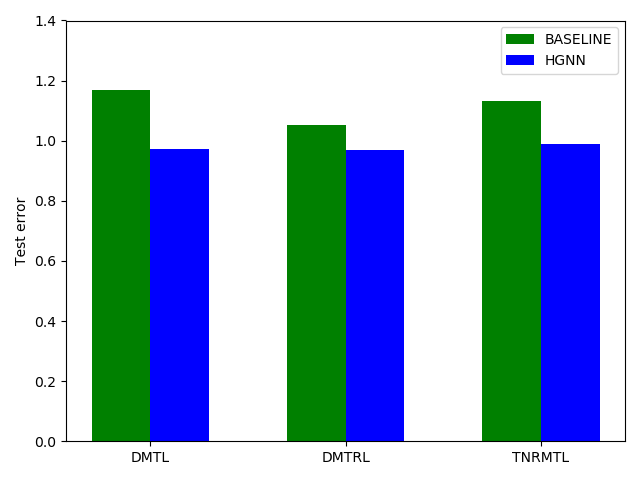}
\vskip -0.1in
\caption{Performance of different models on the SARCOS dataset.}
\label{reg_plot}
\vskip -0.1in
\end{figure}

For regression tasks, the performance measure is the mean square error. The test errors on the SARCOS dataset are shown in Figure \ref{reg_plot} where the training proportion is 70\%. As shown in Figure \ref{reg_plot}, after using the HGNN, the test error of each baseline model has a significant decrease, which demonstrate the effectiveness of augmented features learned in the HGNN method. With other training proportions, we have observed similar phenomena that the use of the HGNN can improve the performance of baseline models, and due to page limit, we did not plot the results in the figures. 

\subsection{Ablation Study}

\begin{figure}[!htbp]
\vskip -0.1in
\centering
\includegraphics[scale=0.35]{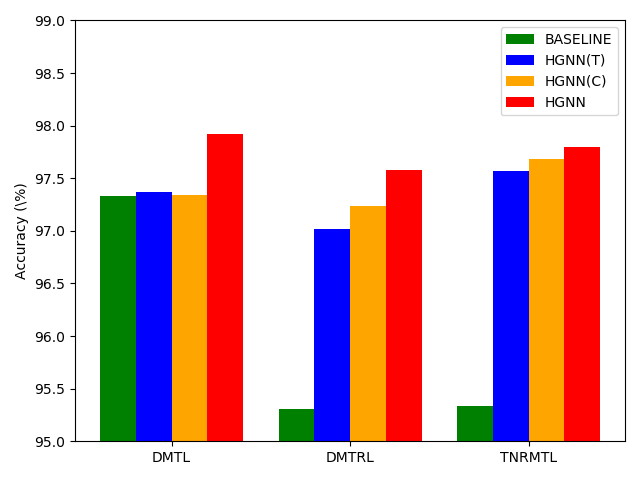}
\vskip -0.1in
\caption{The ablation study on the Office-Caltech-10 dataset.}
\label{ablation_study}
\end{figure}

To study the effectiveness of task embeddings and class embeddings in the HGNN model, we study two variants of HGNN, including HGNN(T) that only augments with the task embedding and HGNN(C) that only augments with the class embedding. The comparison among baseline models, HGNN, variants of HGNN on the Office-Caltech-10 dataset is shown in Figure \ref{ablation_study}. According to the results, we can see that the use of only the class embedding in HGNN(C) or the task embedding in HGNN(T) can improve the performance of baseline models, which shows that augmented features learned in two ways are effective. HGNN(C) seems better than HGNN(T) in this experiment. One reason is that class embeddings may contain more discriminative features for the classification task. Figure \ref{ablation_study} also indicates that using both task embeddings and class embeddings achieves the best performance, which again verifies the usefulness of the HGNN.

\begin{figure*}[!htbp]
\centering
\includegraphics[scale=0.266]{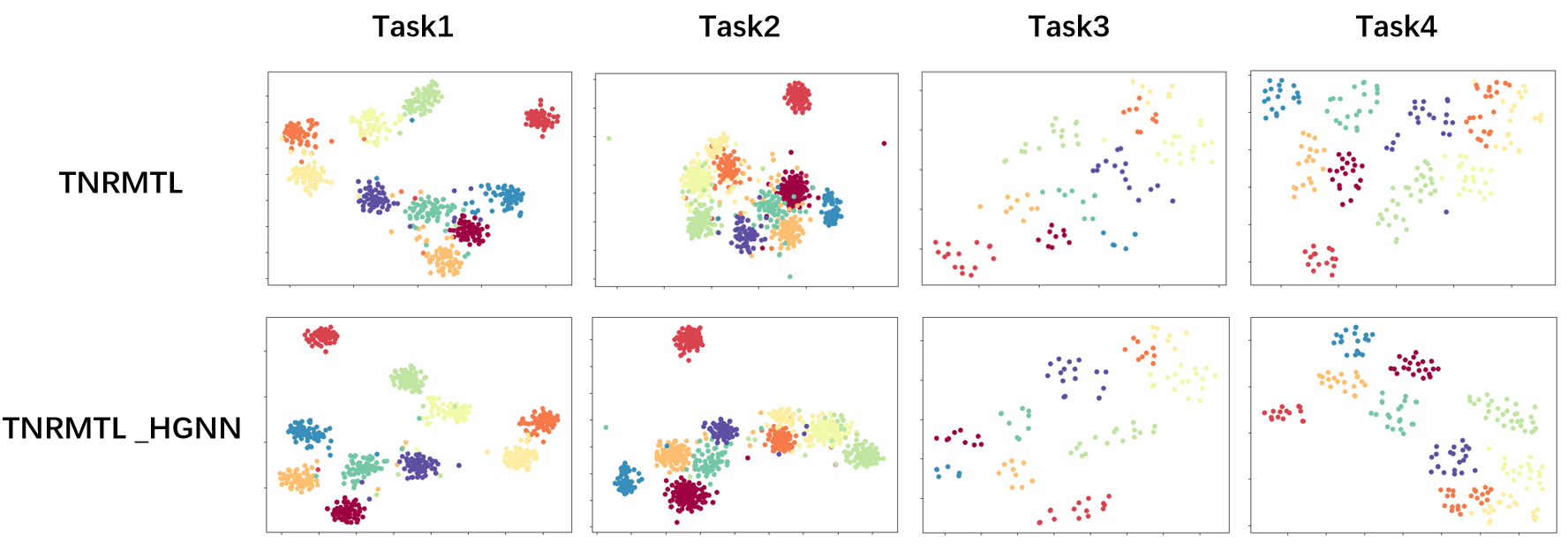}
\caption{The feature visualization by the t-SNE method for the training data in the four tasks on the Office-Caltech-10 dataset. Different markers and different colors are used denote different categories. (Best viewed in color.)}
\label{feature_train}
\end{figure*}

\begin{figure*}[!htbp]
\centering
\includegraphics[scale=0.266]{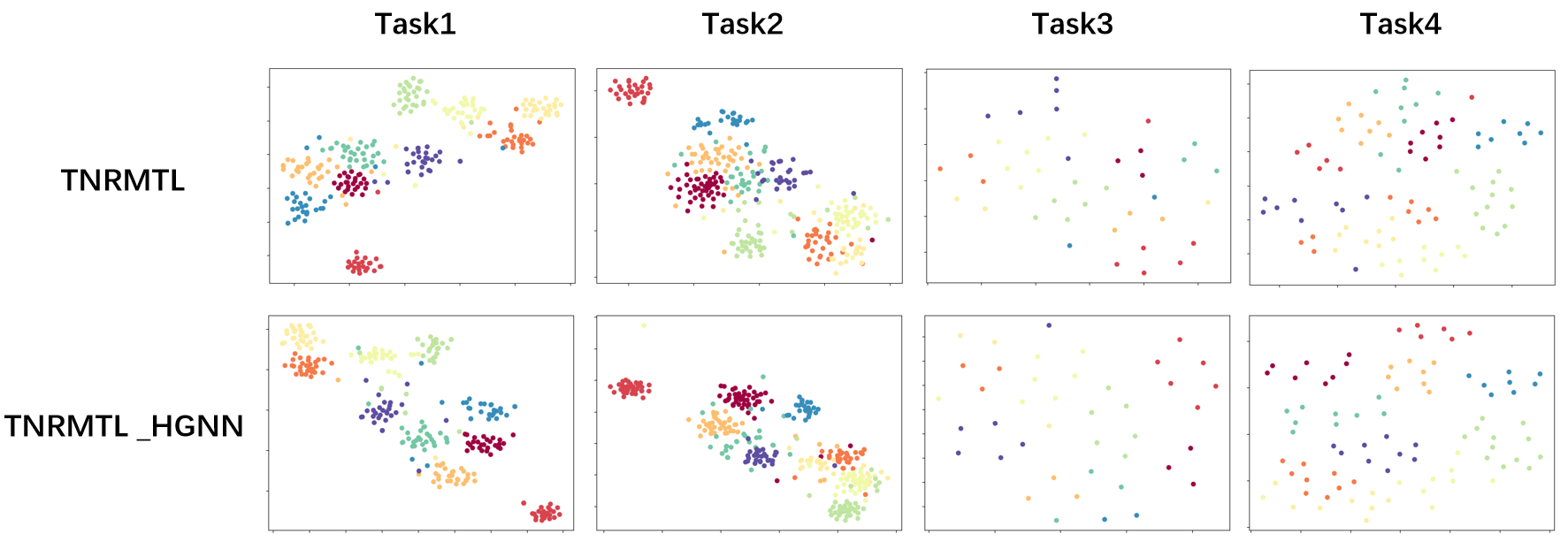}
\caption{The feature visualization by the t-SNE method for the test data in the four tasks on the Office-Caltech-10 dataset. Different markers and different colors are used denote different categories. (Best viewed in color.)}
\label{feature_test}
\end{figure*}

\begin{table*}
\centering
\caption{The classification accuracy (\%) on the ImageCLEF dataset when varying $F'_t$ and fixing $F'_c$ as 8.}
\begin{tabular}{cccccc}
\toprule
 $F'_t$ & 4 & 8 & 16 & 32 & 64 \\
 \midrule
 DMTL\_HGNN & 81.22$\pm$1.27 & \textbf{82.03}$\pm$1.98 & 81.18$\pm$1.43 & 80.41$\pm$0.88 & 79.89$\pm$0.65 \\
 DMTRL\_HGNN & 81.21$\pm$0.80 & \textbf{82.07}$\pm$1.47 & 81.64$\pm$0.77 & 81.15$\pm$2.35 & 81.48$\pm$1.78 \\
 TNRMTL\_HGNN & \textbf{82.35}$\pm$1.66 & 81.11$\pm$0.94 & 82.27$\pm$1.19 & 81.74$\pm$0.91 & 81.20$\pm$0.40 \\
 \bottomrule
 \end{tabular}
 \label{tbl:table1}
\end{table*}

\begin{table*}
\centering
\caption{The classification accuracy (\%) on the ImageCLEF dataset when varying $F'_c$ when $F'_t$ = 8.}
\begin{tabular}{cccccc}
\toprule
 $F'_c$ & 4 & 8 & 16 & 32 & 64 \\
 \midrule
 DMTL\_HGNN & \textbf{82.37}$\pm$1.13 & 82.03$\pm$1.98 & 81.75$\pm$0.56 & 80.71$\pm$1.30 & 81.65$\pm$2.84 \\
 DMTRL\_{HGNN} & 81.73$\pm$1.39 & \textbf{82.07}$\pm$1.47 & 80.40$\pm$1.57 & 81.94$\pm$2.03 & 80.74$\pm$2.16 \\
 TNRMTL\_{HGNN} & 80.07$\pm$3.36 & \textbf{81.11}$\pm$0.94 & 80.64$\pm$1.11 & 80.17$\pm$0.98 & 80.47$\pm$0.80 \\
 \bottomrule
 \end{tabular}
 \label{tbl:table2}
\end{table*}

\subsection{Visualization}

To dive deeper into the learned features, we plot in Figures \ref{feature_train} and \ref{feature_test} the t-SNE embeddings of the feature representations learned for the four tasks on Office-Caltech-10 dataset by TNRMTL and TNRMTL\_{HGNN}, respectively, at the training and testing processes. We observe that the data based on the representation derived by the HGNN model are more separable among classes in each task during either the training process or the testing process. This phenomenon verifies the effectiveness of the augmented features learned in the HGNN to help discriminate data points in different classes of all the tasks.

\subsection{Sensitivity Analysis}

We conduct the sensitivity analysis of the performance with respect to the dimension of task embedding ($F'_t$) and class embedding ($F'_c$), respectively, on the ImageCLEF dataset. The results are shown in Tables \ref{tbl:table1} and \ref{tbl:table2}. According to the results, we can see that
$F'_t = 8$ and $F'_c = 8$ are a good choice in most cases even though in some case, a lower value 4 performs better. When the dimension is not so large (e.g., not large than 32), the performance changes a little, making the choice of the dimension insensitive. However, when using a larger dimension (e.g., 64), the classification accuracy drops significantly, implying that the HGNN prefers a small dimension.

\section{Conclusion}

In this paper, we propose a hierarchical graph neural network (HGNN) to learn augmented features for deep multi-task learning. The proposed HGNN has two levels. In the first level, the intra-task graph neural network is used to learn a powerful representation for each data point in a task by aggregating information from its neighbors in this task. Based on the learned representation, we can learn the task embedding for each task as well as the class embedding if any. The inter-task graph neural network as well inter-class graph neural network is used to update each task embedding and each class embedding. Finally the learned task embedding and class embedding can be used to augment the data representation. Extensive experiments show the effectiveness of the proposed HGNN. In our future work, we are interested in applying the HGNN to other multi-task learning models.



\bibliographystyle{icml2020}
\bibliography{HGNN}

\section*{Appendix}

\subsection*{Proof for Theorem \ref{theorem_training_loss_comparison}}

{\bf Proof}.
By setting the derivation of problems (\ref{obj_analysis_model_1}) and (\ref{obj_analysis_model_2}) to zero, these two problems have close form solutions as
\begin{equation*}
    \begin{aligned}
    \hat{\mathbf{W}}_1 & = (\mathbf{X}\mathbf{X}^\intercal+\lambda \mathbf{I})^{-1}\mathbf{X}\mathbf{Y}^\intercal\\
    \hat{\mathbf{W}}_2 & = (\hat{\mathbf{X}}\hat{\mathbf{X}}^\intercal+\lambda \mathbf{I})^{-1}\hat{\mathbf{X}}\mathbf{Y}^\intercal
    \end{aligned}
\end{equation*}
Then
\begin{equation*}
    \begin{aligned}
    \mathbf{Y}-\hat{\mathbf{W}}_1^\intercal\mathbf{X} & = \mathbf{Y}-\mathbf{Y}\mathbf{X}^\intercal(\mathbf{X}\mathbf{X}^\intercal+\lambda \mathbf{I})^{-\intercal}\mathbf{X}\\&=\mathbf{Y} (\mathbf{I}-\mathbf{X}^\intercal(\mathbf{X}\mathbf{X}^\intercal+\lambda \mathbf{I})^{-\intercal}\mathbf{X})\\ & \triangleq \mathbf{YA}
    \end{aligned}
\end{equation*}
\begin{equation*}
    \begin{aligned}
    \mathbf{Y}-\hat{\mathbf{W}}_2^\intercal\hat{\mathbf{X}} & = \mathbf{Y}-\mathbf{Y}\hat{\mathbf{X}}^\intercal(\hat{\mathbf{X}}\hat{\mathbf{X}}^\intercal+\lambda \mathbf{I})^{-\intercal}\hat{\mathbf{X}} \\&= \mathbf{Y}(\mathbf{I}-\hat{\mathbf{X}}^\intercal(\hat{\mathbf{X}}\hat{\mathbf{X}}^\intercal+\lambda \mathbf{I})^{-\intercal}\hat{\mathbf{X}})\\ & \triangleq \mathbf{YB}.
    \end{aligned}
\end{equation*}
It is easy to show
\begin{equation*}
    (\mathbf{X}\mathbf{X}^\intercal+\lambda \mathbf{I})^{\intercal}\mathbf{X} = \mathbf{X}(\mathbf{X}^\intercal\mathbf{X}+\lambda \mathbf{I})^{\intercal}.
\end{equation*}
By left-multiplying by $(\mathbf{X}\mathbf{X}^\intercal+\lambda \mathbf{I})^{-1}$ and right-multiplying by $(\mathbf{X}^\intercal\mathbf{X}+\lambda \mathbf{I})^{-1}$, we can get
\begin{equation*}
    \mathbf{X}(\mathbf{X}^\intercal\mathbf{X}+\lambda \mathbf{I})^{-\intercal} = (\mathbf{X}\mathbf{X}^\intercal+\lambda \mathbf{I})^{-\intercal}\mathbf{X}.
\end{equation*}
So $\mathbf{A}$ can be simplified as
\begin{equation*}
    \begin{aligned}
    \mathbf{A} & = \mathbf{I}-\mathbf{X}^\intercal(\mathbf{X}\mathbf{X}^\intercal+\lambda \mathbf{I})^{-\intercal}\mathbf{X}\\
      & = \mathbf{I} - \mathbf{X}^\intercal\mathbf{X}(\mathbf{X}^\intercal\mathbf{X}+\lambda \mathbf{I})^{-\intercal}\\
      & = \mathbf{I} - (\mathbf{X}^\intercal\mathbf{X} +\lambda \mathbf{I} -\lambda \mathbf{I})(\mathbf{X}^\intercal\mathbf{X}+\lambda \mathbf{I})^{-1}\\
      & = \lambda(\mathbf{X}^\intercal\mathbf{X}+\lambda \mathbf{I})^{-1}
    \end{aligned}
\end{equation*}
In order to prove
\begin{equation*}
    \begin{aligned}
    &\|\mathbf{Y} - \hat{\mathbf{W}}_1^\intercal\mathbf{X}\|_2^2-\|\mathbf{Y}-\hat{\mathbf{W}}_2^\intercal\hat{\mathbf{X}}\|_2^2\\
    =&\|\mathbf{YA}\|_2^2-\|\mathbf{YB}\|_2^2\\
    \geq & 0,
    \end{aligned}
\end{equation*}
We need to require
\begin{equation}
   \mathbf{A}^\intercal\mathbf{A}\succeq \mathbf{B}^\intercal\mathbf{B},\label{theorem_1_proof_equation_1}
\end{equation}
where $\mathbf{A} = \lambda(\mathbf{X}^\intercal\mathbf{X}+\lambda \mathbf{I})^{-1},\mathbf{B} = \lambda(\hat{\mathbf{X}}^\intercal\hat{\mathbf{X}}+\lambda \mathbf{I})^{-1}$.
Note that
\begin{equation*}
    \begin{aligned}
    \mathbf{B} &= \lambda((\mathbf{X}^\intercal,\mathbf{E}^\intercal)(\mathbf{X}^\intercal,\mathbf{E}^\intercal)^\intercal+\lambda \mathbf{I})^{-1}\\&=\lambda(\mathbf{X}^\intercal\mathbf{X}+\mathbf{E}^\intercal\mathbf{E}+\lambda \mathbf{I})^{-1}
    \end{aligned}
\end{equation*}
Eq. (\ref{theorem_1_proof_equation_1}) is equivalent to
\begin{equation*}
    \begin{aligned}
    &\big((\mathbf{X}^\intercal\mathbf{X}+\lambda \mathbf{I})(\mathbf{X}^\intercal\mathbf{X}+\lambda \mathbf{I})\big)^{-1}\\ &\succeq \big((\mathbf{X}^\intercal\mathbf{X}+\mathbf{E}^\intercal\mathbf{E}+\lambda \mathbf{I})(\mathbf{X}^\intercal\mathbf{X}+\mathbf{E}^\intercal\mathbf{E}+\lambda \mathbf{I})\big)^{-1},
    \end{aligned}
\end{equation*}
which is also equivalent to
\begin{equation*}
    \begin{aligned}
    &(\mathbf{X}^\intercal\mathbf{X}+\mathbf{E}^\intercal\mathbf{E}+\lambda \mathbf{I})(\mathbf{X}^\intercal\mathbf{X}+\mathbf{E}^\intercal\mathbf{E}+\lambda \mathbf{I})\\ &\succeq (\mathbf{X}^\intercal\mathbf{X}+\lambda \mathbf{I})(\mathbf{X}^\intercal\mathbf{X}+\lambda \mathbf{I}).
    \end{aligned}
\end{equation*}
So we require
\begin{equation*}
    \begin{aligned}
    \mathbf{X}^\intercal\mathbf{X}\mathbf{E}^\intercal\mathbf{E}+\mathbf{E}^\intercal\mathbf{E}\mathbf{X}^\intercal\mathbf{X}+2\lambda \mathbf{E}^\intercal\mathbf{E} +\mathbf{E}^\intercal\mathbf{E}\mathbf{E}^\intercal\mathbf{E} \succeq 0
    \end{aligned}
\end{equation*}
which is the condition.
\hfill $\Box$

\subsection*{Proof for Theorem \ref{theorem_generalization_bound}}

Before proving Theorem \ref{theorem_generalization_bound}, we introduce a lemma from \cite{NIPS2008_3510}.

\begin{lemma}\label{lemma_generalization_bound}
If we assume $\|\mathbf{x}\|\leq X$, $F(\mathbf{W})\leq W_*$. For any Lipschitz loss function $\ell$, with Lipschitz constant $L_\ell$. For any $\delta>0$ and with probability at least $1-\delta$ over the sample:
Let $\mathcal{W}$ be as in the $L_p$ \textbf{norms} example, that is $F(\mathbf{W})=\|\mathbf{W}\|_q^2$, where $ \frac{1}{p}+\frac{1}{q}=1$. For all $\mathbf{w}\in \mathcal{W}$, we have
\begin{equation*}
    \mathcal{L}(\mathbf{w})\leq \hat{\mathcal{L}}(\mathbf{w})+2L_\ell X W_*\sqrt{\frac{p-1}{n}}+L_\ell X W_*\sqrt{\frac{\log(1/\delta)}{2n}}.
\end{equation*}
\end{lemma}

Then we can give a proof of Theorem \ref{theorem_generalization_bound}.

{\bf Proof}. Since the loss function has a Lipschitz constant 2, according to Lemma \ref{lemma_generalization_bound}, we let $p=q=2$ and then we can reach the conclusion.
\hfill $\Box$

\end{document}